\documentclass{article}
\usepackage{spconf,amsmath,amsfonts,graphicx}
\usepackage[utf8]{inputenc} 
\usepackage[T1]{fontenc}    
\usepackage{url}            
\usepackage{booktabs}       
\usepackage{nicefrac}       
\usepackage{microtype}      
\usepackage{xcolor}         
\usepackage{multirow}
\usepackage{subcaption}
\usepackage{cuted}
\usepackage{hyperref}       
\usepackage{enumitem}

\title{VLM Fine-Tuning for End-to-End Combinatorial Optimization}
\begin{document}
%
\maketitle
\begin{abstract}
Large language models (LLMs) have provided a unified interface for end-to-end
combinatorial optimization (CO), but textual serialization alone may obscure
spatial and relational structures that are important for generating effective CO solutions.
This paper presents a general-purpose vision-language solver that augments textual instance descriptions with input-derived visual representations. A single vision-language model (VLM) is applied across different CO tasks and trained using
supervised fine-tuning followed by verifier-guided reinforcement learning. While the
visual inputs contain no gold solutions or solution-derived information, our
experiments show that the VLM generally improves solution quality over its
text-only counterpart, with particularly clear gains on more complex CO problems such as CVRP and JSSP. 
The advantage of visual information is more pronounced at large problem scales.
\end{abstract}
\section{Introduction}
\label{sec:intro}

Combinatorial optimization (CO) problems arise in routing, scheduling,
allocation, and network design. Deep learning-based CO solvers have achieved strong results,
but they commonly rely on problem-specific architectures and training
procedures \cite{berto2023rl4co,reijnen2026job}. Recent work
has explored large language models (LLMs) for automatic heuristic design
\cite{eoh,ye2024reevo} and optimization formulation \cite{jiang2025llmopt,jiang2025droc}. However, these approaches still require domain expertise in designing algorithmic templates or using optimization solvers. In contrast, end-to-end LLM solvers formulate CO problems directly in natural language and generate solutions, 
providing a more accessible and unified solving paradigm while reducing the dependence on specialized domain knowledge \cite{jiang2026large,liu2026hard}. While end-to-end
solvers provide a flexible language interface, purely textual serialization 
makes spatial and relational structures, such as graph connectivity and object relationships, difficult to perceive.

This limitation motivates us to investigate whether visual information can
improve LLM-based CO solving. Specifically, we represent each CO instance with a textual description and an input-derived image that exposes its
problem-relevant structure. A single vision-language model (VLM) is trained across different CO problems through
supervised fine-tuning (SFT) and verifier-guided reinforcement learning (RL).
The visualization and verification procedures are tailored to the structure of each problem, while the underlying VLM and training framework remain shared. Our main contributions are summarized below: 1) We introduce a multimodal representation for heterogeneous CO
    problems, combining textual instance descriptions with associated images, without incorporating any solution-derived information. 2) As a first attempt, we train a shared VLM for CO problems using SFT and RL to translate joint vision-language inputs into textual solution descriptions. 3) Extensive experiments demonstrate that visual information improves end-to-end CO solving, yielding substantial gains over the text-only solver and stronger generalization as problem size increases.


\begin{figure*}[t]
    \centering
    \includegraphics[
        width=\textwidth,
        keepaspectratio
    ]{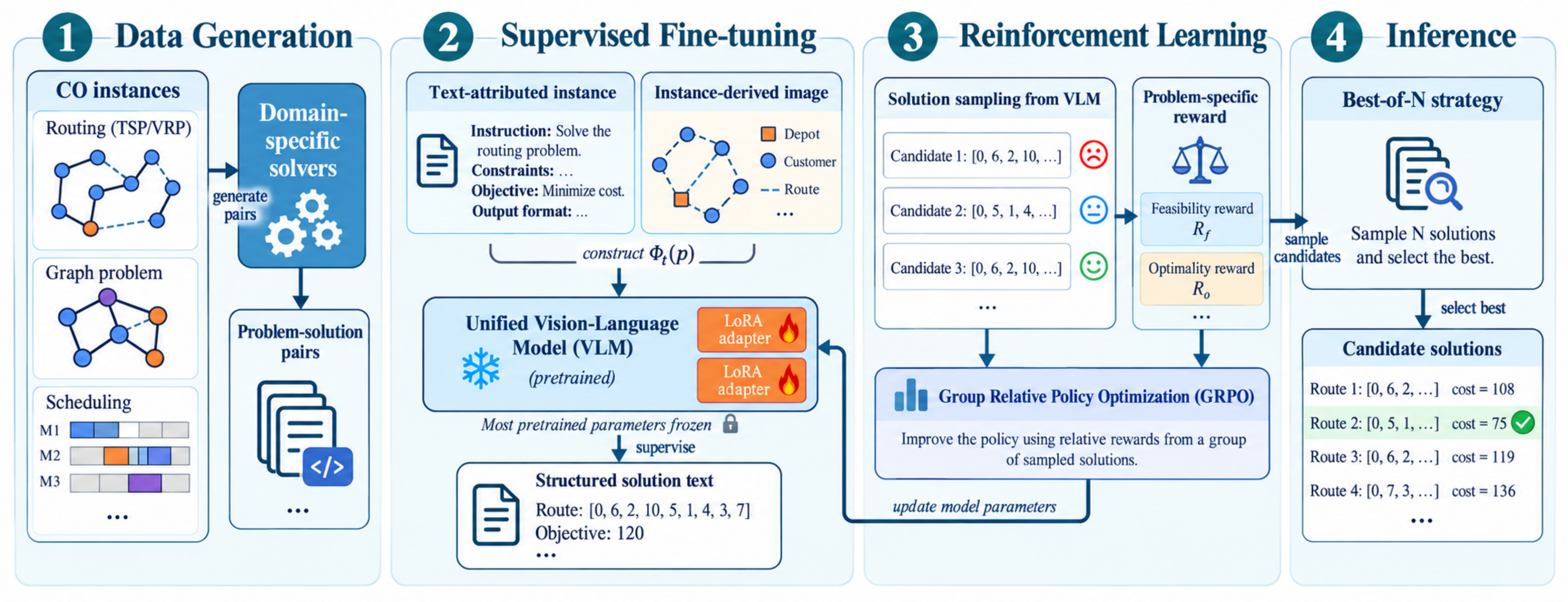}
    \caption{Overview of the proposed vision-language CO solver.}
    \label{fig:framework}
\end{figure*}

\section{Problem Statement}
\label{sec:problem_statement}

Let $\mathcal{T}$ denote a set of CO problems and $\mathcal{P}_t$ the
instance space of problem $t\in\mathcal{T}$. Each instance
$p\in\mathcal{P}_t$ is represented by $(X_p,f_p,C_p)$. $X_p$ is
the solution space, $f_p:X_p\rightarrow\mathbb{R}$ is the objective
function, and $C_p=\{c_{p,1},\ldots,c_{p,m_p}\}$ is the constraint set,
with $c_{p,i}:X_p\rightarrow\{0,1\}$ denoting whether $c_{p,i}$ is satisfied. The feasible solution space  for problem $t$ is $X_p^{\mathrm{F}}
=
\left\{
x\in X_p
\mid
c_{p,i}(x)=1,\ \forall c_{p,i}\in C_p
\right\}$.
The goal is to find a solution in $X_p^{\mathrm{F}}$ that minimizes or
maximizes $f_p$, depending on the task. Unlike text-only CO solvers \cite{jiang2026large} using pure texts, we represent each instance using both textual
and visual information. Let $\mathcal{S}$ and $\mathcal{I}$ denote the text
and image spaces. A problem-specific textual serialization
$\phi_t^{\mathrm{text}}\in \mathcal{S}$ preserves exact instance information (e.g., parameters and constraints),
while an image $\phi_t^{\mathrm{img}}\in \mathcal{I}$ provides complementary
structural information. The multimodal input is defined as $\Phi_t(p)
=
(
\phi_t^{\mathrm{text}}(p),
\phi_t^{\mathrm{img}}(p)
)$.
The text and image are generated solely from the input instance, without leakage of any solution-derived information.

A single VLM policy $\pi_\theta$ is shared across all problems. Given
$\Phi_t(p)$, 
it generates a textual response $\hat{y}_p
\sim
\pi_\theta(\cdot\mid\Phi_t(p))$, which can be  easily converted into a numerical solution by a parser $\psi_t$, i.e., $\hat{x}_p=\psi_t(\hat{y}_p)$.
A verifier independently evaluates the parsed solution using
the original instance:
\begin{equation}
(v_p,\hat{f}_p)=\mathcal{V}_t(\hat{x}_p,p),
\label{eq:task_verifier}
\end{equation}
where $v_p\in\{0,1\}$ indicates feasibility and
$\hat{f}_p=f_p(\hat{x}_p)$ is the objective value of the solution. Let $\mathcal{D}_s$ denote the evaluation set and
$f_p^{\mathrm{ref}}$ be the optimal or best-known reference objective. For
a feasible solution, its relative gap is:
\begin{equation}
g_p
=
\begin{cases}
\hat{f}_p-f_p^{\mathrm{ref}}/{|f_p^{\mathrm{ref}}|},
& \text{for minimization},\\[7pt]
f_p^{\mathrm{ref}}-\hat{f}_p/{|f_p^{\mathrm{ref}}|},
& \text{for maximization}.
\end{cases}
\label{eq:relative_gap}
\end{equation}
Following the literature, we report the feasibility rate and the average gap over feasible solutions:
\begin{equation}
M_f
=
\frac{1}{|\mathcal{D}_s|}
\sum_{(t,p)\in\mathcal{D}_s}v_p,
\qquad
M_o
=
\frac{1}{|\mathcal{D}_s^{\mathrm{F}}|}
\sum_{(t,p)\in\mathcal{D}_s^{\mathrm{F}}}g_p,
\label{eq:evaluation_metrics}
\end{equation}
where $\mathcal{D}_s^{\mathrm{F}}$ contains the instances for which $v_p=1$.

\section{The Method}
\label{sec:method}

Figure~\ref{fig:framework} illustrates our method. The same VLM
processes heterogeneous CO problems, while problem-specific components are  limited to input construction and solution verification.


\begin{figure}[!t]
    \centering
    \begin{minipage}[t]{0.23\textwidth}
        \centering
        \includegraphics[width=0.9\linewidth]{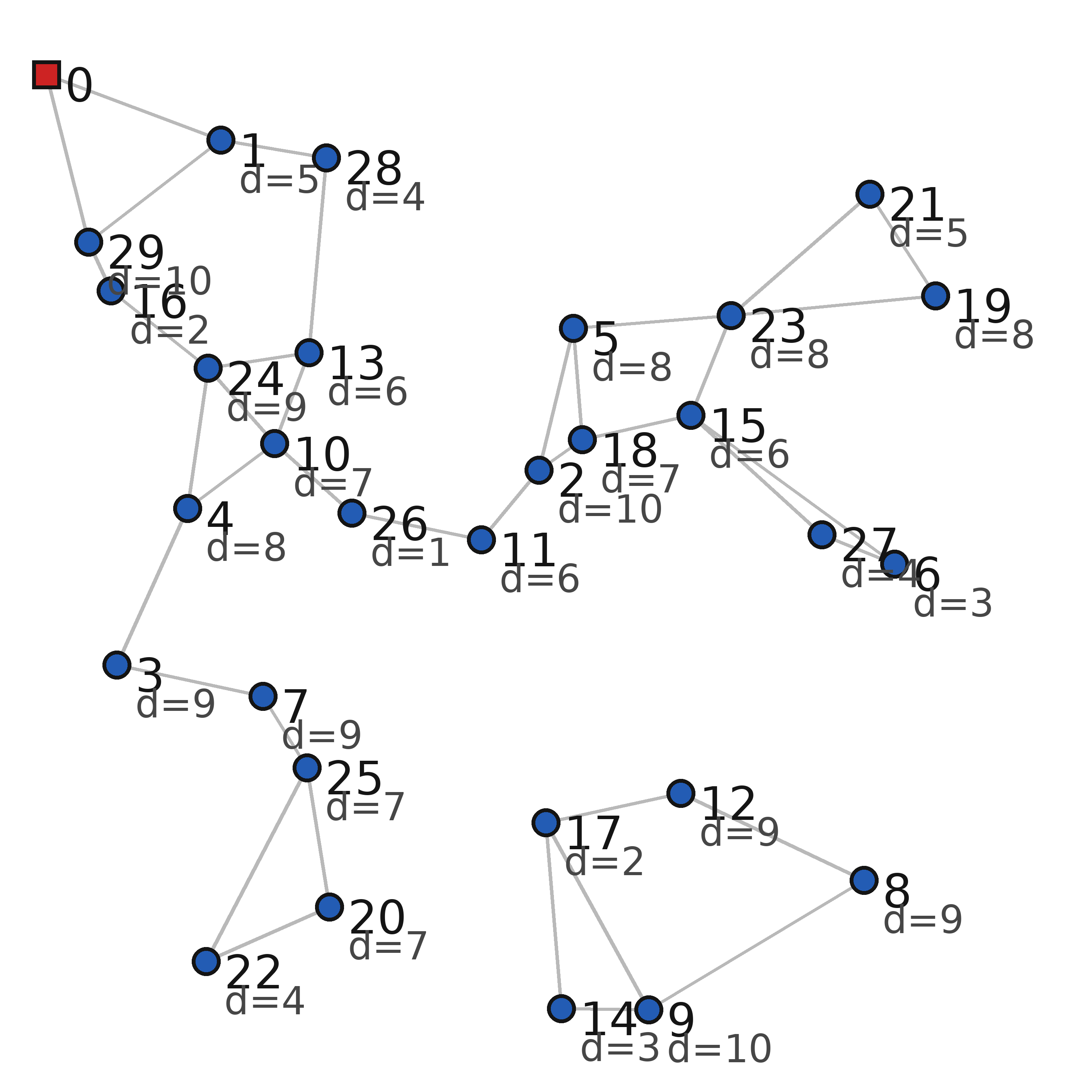}
    \end{minipage}
    \hspace{4pt}
    \begin{minipage}[t]{0.23\textwidth}
        \centering
        \includegraphics[width=0.9\linewidth]{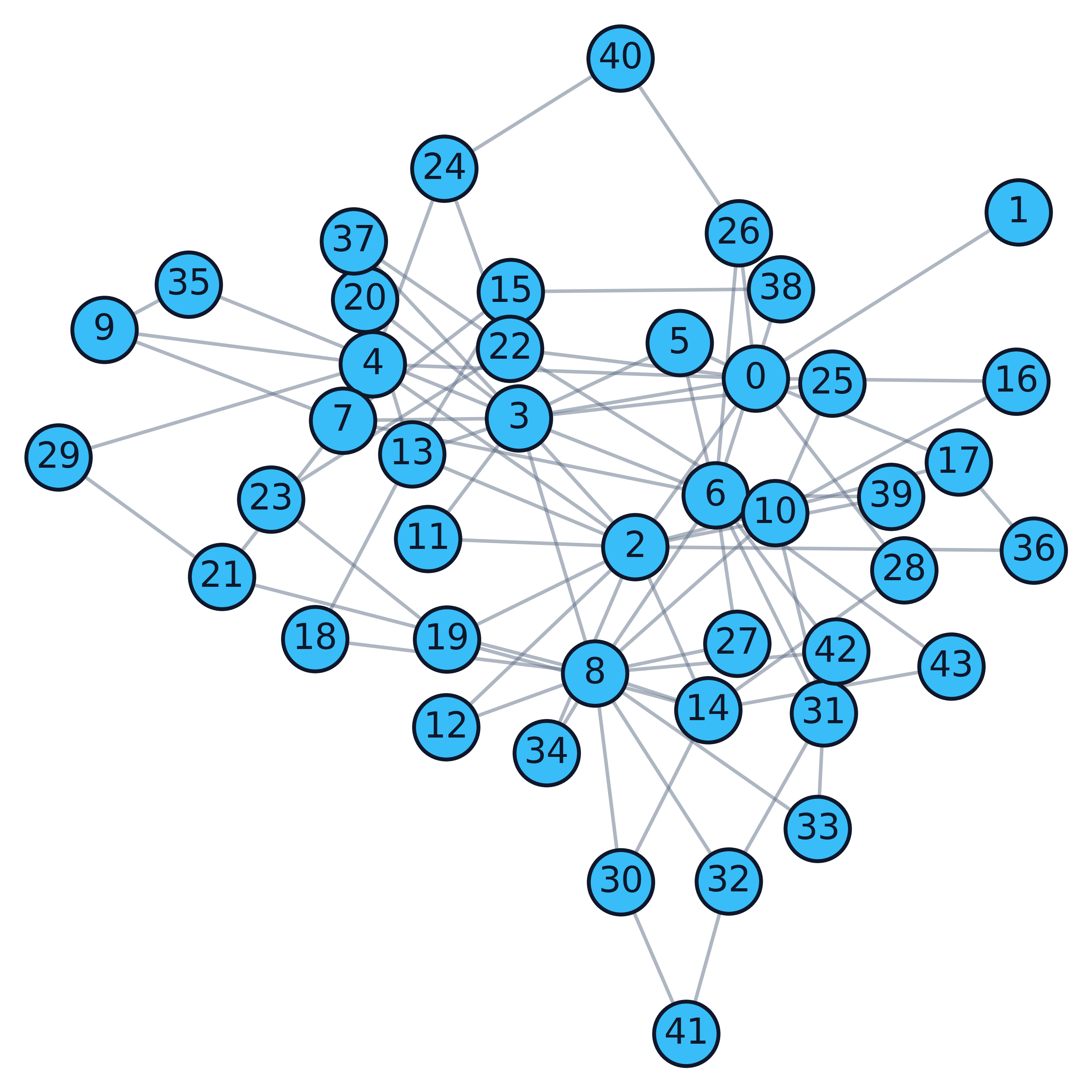}
    \end{minipage}

    \vspace{-4pt}
    \caption{Visual Representation of CVRP (left) and MIS (right).}
    \vspace{-14pt} 
    \label{fig:app_cvrp_visuals}
\end{figure}

\subsection{Visual Representation}
\label{sec:vision_formats}

We design problem-adapted images to reveal structural relationships that are hard to capture from serialized text alone. For routing problems (e.g., TSP, CVRP, and OP), we adopt a unified node–link visualization paradigm. Taking TSP as an example, each node in the image denotes one customer. Its location is represented by two-dimensional coordinates, such that the spatial distance between two nodes is proportional to their distance. Node 0 is rendered as a red square to mark the designated starting customer, while all other cities are drawn as blue circles with labels indicating customer IDs. Each edge in the image connects a customer to its two nearest neighboring cities.
CVRP and OP follow the same node-link construction, with additional node attributes. 
For an example in Fig.~\ref{fig:app_cvrp_visuals},  
customer labels contain IDs and demands in CVRP.
For graph problems such as MIS and MVC, the node–link image directly visualizes the input graph. Each labeled blue circle denotes a node, and each gray line is an undirected edge to represent the graph adjacency. 
For scheduling problems, we represent them by a graph. Taking JSSP as the typical case,  nodes represent operations with features encoding processing times, while edge features distinguish job precedence. 

For all problems, the textual input remains the authoritative source for exact numerical values in instances, which follows the natural-language instance paradigm widely adopted in existing literature. More details of problem definition and textual description can be found in \cite{jiang2026large}.




\subsection{Supervised Fine-tuning}
\label{sec:sft}

Given a multimodal input $\Phi_t(p)$ and its target textual response
$y_p^*=(y_{p,1}^*,\ldots,y_{p,L_p}^*)$, the SFT objective is the standard
autoregressive language-modeling loss:
\begin{equation}
\mathcal{L}_{\mathrm{SFT}}(\theta)
=
-
\mathbb{E}_{(t,p)}
\sum_{k=1}^{L_p}
\log
\pi_\theta
\left(
y_{p,k}^*
\mid
\Phi_t(p),y_{p,<k}^*
\right),
\label{eq:sft_loss}
\end{equation}
which is applied only to tokens in the target response, while the
instruction, textual instance description, image tokens, and output
format prompt are treated as conditioning information. Masking these information encourages the VLM to learn the generation of solutions rather than reproducing input prompt. To reduce computational cost, we apply Low-Rank Adaptation (LoRA)
\cite{hu2022lora} while freezing the pretrained VLM parameters.
The resulting SFT-trained model provides the initial policy for the subsequent RL.

\subsection{Reinforcement Learning}
\label{sec:rl}

Despite SFT teaching the VLM to imitate high-quality solutions, it does not
guarantee constraint satisfaction. Therefore we further update the policy using constraint-verified RL. For a multimodal instance $\Phi_t(p)$, the current policy samples a
group of $G$ candidate textual responses: $\hat{y}_{p,1},\ldots,\hat{y}_{p,G}
\sim
\pi_{\theta_{\mathrm{old}}}
\left(
\cdot\mid\Phi_t(p)
\right)$,
with each response parsed and independently
evaluated on the instance. After that, the RL reward is designed by a
feasibility component and an optimality component:
\begin{equation}
R_{p,i}
=
\lambda_f R_{f,t}(\hat{y}_{p,i},p)
+
\lambda_o R_{o,t}(\hat{y}_{p,i},p),
\label{eq:combined_reward}
\end{equation}
where $R_{f,t}$ measures constraint satisfaction,
and $R_{o,t}$ measures the quality in terms of the objective value. 
$R_{o,t}$ is assigned only when the generated candidate solution satisfies the required constraints. 
Concretely, the feasibility verifier evaluates each problem-specific constraint against the original instance. $R_{f,t}$ is computed as a weighted sum of the satisfied constraints. If the generated solution is feasible, the objective verifier computes its objective value from the parsed solution.

We use Group Relative Policy Optimization (GRPO) \cite{guo2025deepseek} to update the VLM. The advantage of the
$i$-th response is computed relative to the other responses in the same
group:
\begin{equation}
A_{p,i}
=
\frac{
    R_{p,i}-\mu_p
}{
    \sigma_p+\delta
},
\label{eq:group_advantage}
\end{equation}
where $\mu_p$ and $\sigma_p$ are the mean and standard deviation of the group
rewards, and $\delta$ is a numerical stability constant. The policy is
optimized using the clipped objective:
\begin{equation}
\begin{aligned}
J_{\mathrm{GRPO}}(\theta)
=&{}\mathbb{E}\Big[
\frac{1}{G}\sum_{i=1}^{G}
\min\Big\{r_{p,i}A_{p,i}, \operatorname{clip}\big(r_{p,i},\\
&1-\epsilon,1+\epsilon\big)A_{p,i}\Big\}
-\beta D_{\mathrm{KL}}
\big(\pi_\theta\,\Vert\,\pi_{\mathrm{ref}}\big)
\Big],
\end{aligned}
\label{eq:grpo_objective}
\end{equation}
where
\[
r_{p,i}
=
\frac{
\pi_\theta(\hat{y}_{p,i}\mid\Phi_t(p))
}{
\pi_{\theta_{\mathrm{old}}}(\hat{y}_{p,i}\mid\Phi_t(p))
}
\]
is the policy ratio, $\epsilon$ is the clipping parameter, and $\beta$
controls the divergence from the reference policy.

Through this feasibility-and-optimality-aware RL stage, the VLM is
encouraged to preserve the structured generation ability learned during
SFT while reducing constraint violations and improving the quality of
feasible solutions.

\begin{table*}[t]
\centering
\captionof{table}{Evaluation of feasibility (Feas.$\uparrow$), relative gap (Gap$\downarrow$), and average time for different methods on the seven CO problems.}
\setlength{\tabcolsep}{0.7pt}
\renewcommand{\arraystretch}{0.62}
\resizebox{\textwidth}{!}{%
\begin{tabular}{l *{14}{c} r}
\hline
\addlinespace[1pt]
\multirow{2}{*}{Method} & \multicolumn{2}{c}{\textbf{TSP}} & \multicolumn{2}{c}{\textbf{OP}} & \multicolumn{2}{c}{\textbf{CVRP}} & \multicolumn{2}{c}{\textbf{MIS}} & \multicolumn{2}{c}{\textbf{MVC}} & \multicolumn{2}{c}{\textbf{PFSP}} & \multicolumn{2}{c}{\textbf{JSSP}} & \multirow{2}{*}{\textbf{ Time}} \\
\cmidrule(lr){2-3} \cmidrule(lr){4-5} \cmidrule(lr){6-7} \cmidrule(lr){8-9} \cmidrule(lr){10-11} \cmidrule(lr){12-13} \cmidrule(lr){14-15}
& \textbf{Feas.} & \textbf{Gap} & \textbf{Feas.} & \textbf{Gap} & \textbf{Feas.} & \textbf{Gap} & \textbf{Feas.} & \textbf{Gap} & \textbf{Feas.} & \textbf{Gap} & \textbf{Feas.} & \textbf{Gap} & \textbf{Feas.} & \textbf{Gap} & \\
\midrule
\multicolumn{16}{l}{\textbf{General-purpose Language Models}} \\
\addlinespace[1pt]
\hline
\addlinespace[1pt]
GPT-4o & 39\% & 33.79\%$_{\pm16.6}$ & 59\% & 55.19\%$_{\pm15.7}$ & 15\% & 76.62\%$_{\pm7.9}$ & 8\% & 11.70\%$_{\pm11.8}$ & 6\% & 16.67\%$_{\pm7.0}$ & 88\% & 20.57\%$_{\pm9.2}$ & 7\% & 97.85\%$_{\pm23.7}$ & 5.3s \\
Claude-Sonnet & 66\% & 24.53\%$_{\pm10.7}$ & 49\% & 34.62\%$_{\pm14.1}$ & 30\% & 38.34\%$_{\pm15.9}$ & 13\% & 12.51\%$_{\pm12.5}$ & 2\% & 6.25\%$_{\pm6.3}$ & 100\% & 18.42\%$_{\pm8.9}$ & 10\% & 90.00\%$_{\pm21.6}$ & 5.4s \\
DeepSeek-V3 & 73\% & 35.75\%$_{\pm15.4}$ & 50\% & 46.10\%$_{\pm13.4}$ & 21\% & 58.22\%$_{\pm26.8}$ & 5\% & 12.05\%$_{\pm12.9}$ & 15\% & 37.15\%$_{\pm24.8}$ & 58\% & 20.81\%$_{\pm9.4}$ & 52\% & 103.19\%$_{\pm26.9}$ & 26.4s \\
Llama3.3-70B & 50\% & 69.08\%$_{\pm31.4}$ & 27\% & 48.98\%$_{\pm14.6}$ & 31\% & 97.31\%$_{\pm69.3}$ & 8\% & 37.12\%$_{\pm29.5}$ & 20\% & 22.86\%$_{\pm13.6}$ & 98\% & 21.97\%$_{\pm8.4}$ & 29\% & 105.01\%$_{\pm24.5}$ & 2.1s \\
Qwen2.5-72B & 20\% & 36.89\%$_{\pm34.6}$ & 32\% & 49.36\%$_{\pm16.6}$ & 61\% & 180.91\%$_{\pm105}$ & 14\% & 29.56\%$_{\pm16.2}$ & 5\% & 63.20\%$_{\pm30.4}$ & 98\% & 21.13\%$_{\pm8.2}$ & 53\% & 103.90\%$_{\pm61.7}$ & 12.5s \\
\midrule
\multicolumn{16}{l}{\textbf{Reasoning Models}} \\
\hline
\addlinespace[1pt]
GPT-o3-mini & 91\% & 306\%$_{\pm258}$ & 8\% & 43.93\%$_{\pm11.5}$ & 50\% & 139\%$_{\pm39.7}$ & 66\% & 9.23\%$_{\pm8.4}$ & 33\% & 2.98\%$_{\pm5.5}$ & 98\% & 16.97\%$_{\pm9.8}$ & 20\% & 77.86\%$_{\pm37.8}$ & 1.4m \\
GPT-o1 & 54\% & 276\%$_{\pm242}$ & 31\% & 40.90\%$_{\pm16.3}$ & 24\% & 154\%$_{\pm117}$ & 82\% & 8.03\%$_{\pm12.9}$ & 47\% & 3.58\%$_{\pm5.9}$ & 89\% & 14.86\%$_{\pm10.5}$ & 29\% & 81.90\%$_{\pm29.6}$ & 3.2m \\
DeepSeek-R1 & 48\% & 70.99\%$_{\pm23.1}$ & 60\% & 40.54\%$_{\pm13.7}$ & 26\% & 30.46\%$_{\pm18.1}$ & 41\% & 1.60\%$_{\pm3.6}$ & 38\% & 4.17\%$_{\pm6.3}$ & 100\% & 16.65\%$_{\pm8.1}$ & 5\% & 26.29\%$_{\pm8.5}$ & 6.5m \\
\midrule
\multicolumn{16}{l}{\textbf{Prompt Strategies}} \\
\hline
\addlinespace[1pt]
OPRO & 83\% & 35.98\%$_{\pm3.6}$ & 85\% & 53.96\%$_{\pm14.3}$ & 21\% & 37.03\%$_{\pm18.3}$ & 7\% & 5.95\%$_{\pm7.3}$ & 9\% & 41.67\%$_{\pm16.9}$ & 99\% & 18.40\%$_{\pm8.4}$ & 65\% & 83.35\%$_{\pm25.5}$ & 2.1m \\
LMEA & 77\% & 265\%$_{\pm131}$ & 48\% & 66.18\%$_{\pm10.5}$ & 24\% & 61.24\%$_{\pm19.0}$ & 5\% & 25.0\%$_{\pm14.2}$ & 13\% & 34.22\%$_{\pm16.3}$ & 98\% & 14.31\%$_{\pm7.1}$ & 44\% & 83.19\%$_{\pm23.9}$ & 5.3m \\
PHP & 84\% & 33.84\%$_{\pm14.6}$ & 43\% & 36.08\%$_{\pm15.1}$ & 33\% & 58.11\%$_{\pm26.4}$ & 5\% & 11.67\%$_{\pm9.1}$ & 13\% & 19.84\%$_{\pm10.01}$ & 92\% & 17.23\%$_{\pm8.1}$ & 56\% & 104.04\%$_{\pm29.2}$ & 1.6m \\
SGE & 98\% & 29.66\%$_{\pm43.3}$ & 93\% & 24.49\%$_{\pm38.4}$ & 84\% & 36.14\%$_{\pm59.2}$ & 92\% & 3.62\%$_{\pm7.6}$ & 94\% & 3.83\%$_{\pm7.4}$ & 95\% & 4.48\%$_{\pm7.4}$ & 87\% & 38.58\%$_{\pm49.2}$ & 3.6m \\
\midrule
\multicolumn{16}{l}{\textbf{Text-only End-to-End Solver}} \\
\hline
\addlinespace[1pt]
SFT & 89\% & 2.30\%$_{\pm1.9}$ & 54\% & 2.32\%$_{\pm2.6}$ & 59\% & 6.02\%$_{\pm3.9}$ & 80\% & 1.71\%$_{\pm3.9}$ & 98\% & 2.41\%$_{\pm3.3}$ & 100\% & 2.22\%$_{\pm1.9}$ & 100\% & 11.01\%$_{\pm7.9}$ & 5.6s \\
SFT+RL & 91\% & 2.32\%$_{\pm2.2}$ & 92\% & 4.25\%$_{\pm2.9}$ & 80\% & 8.27\%$_{\pm5.6}$ & 83\% & 1.34\%$_{\pm3.3}$ & 98\% & 2.39\%$_{\pm3.2}$ & 100\% & 2.12\%$_{\pm1.8}$ & 100\% & 10.94\%$_{\pm7.3}$ & 5.6s \\
SFT+RL+BoN & \textbf{ 100\%} & 1.07\%$_{\pm0.9}$ & 100\% & 1.85\%$_{\pm1.7}$ & \textbf{ 100\%} & 4.53\%$_{\pm3.5}$ & 94\% & \textbf{ 1.04\%$_{\pm3.4}$} & 100\% & 1.29\%$_{\pm2.2}$ & 100\% & 1.03\%$_{\pm1.1}$ & 100\% & 8.20\%$_{\pm6.3}$ & 9.8s \\
\midrule
\multicolumn{16}{l}{\textbf{Vision-Language Solver (Ours)}} \\
\hline
\addlinespace[1pt]
SFT & 78\% & 2.15\%$_{\pm1.3}$ & 63\% & 1.67\%$_{\pm1.2}$ & 48\% & 4.43\%$_{\pm4.8}$ & 86\% & 1.50\%$_{\pm3.7}$ & 94\% & 2.13\%$_{\pm3.4}$ & 100\% & 2.13\%$_{\pm1.8}$ & 97\% & 8.31\%$_{\pm6.18}$ & 7.3s\\
SFT+RL & 98\% & 2.40\%$_{\pm2.6}$ & 97\% & 2.12\%$_{\pm1.8}$ & 77\% & 4.74\%$_{\pm5.1}$ &95\% &1.67\%$_{\pm3.5}$ & 95\%&1.80\%$_{\pm2.5}$ & 100\% & 2.49\%$_{\pm2.4}$ & 97\% & 8.08\%$_{\pm6.4}$ & 7.3s\\
SFT+RL+BoN & \textbf{ 100\%} & \textbf{ 0.91\%$_{\pm1.1}$} & \textbf{ 100\%} & \textbf{ 1.13\%$_{\pm0.9}$} & 98\% & \textbf{ 3.58\%$_{\pm3.0}$} &\textbf{ 97\%} &1.14\%$_{\pm3.1}$ &\textbf{ 100\%} & \textbf{ 1.11\%$_{\pm2.3}$} & \textbf{ 100\%} & \textbf{ 1.03\%$_{\pm1.7}$} & \textbf{ 100\%} & \textbf{ 5.85\%$_{\pm5.11}$} & 33.8s\\
\hline
\end{tabular}%
}
\label{tab:model_summary_vlm}
\end{table*}

\begin{table*}[t]
\centering
\captionof{table}{Performance comparison across CO problems with different instance scales.}
\renewcommand{\arraystretch}{0.60}
\setlength{\tabcolsep}{5pt}
\scriptsize  
\begin{tabular}{ll|cccc|cccc|cccc}
    \hline
    \multirow{6}{*}[-4ex]{\rotatebox[origin=c]{90}{\scriptsize{\emph{TSP}}}} & \multirow{2}{*}{Method} & \multicolumn{4}{c}{\textbf{Small instances}} & \multicolumn{4}{c}{\textbf{Medium instances}} & \multicolumn{4}{c}{\textbf{Large instances}} \\
    & & Gap & Gap@1 & Gap@5 & Gap@10 & Gap & Gap@1 & Gap@5 & Gap@10 & Gap & Gap@1 & Gap@5 & Gap@10 \\
    \midrule
     & OR-Tools  & 0.82\% & 76\% & 96\% & 99\% & 2.59\% & 28\% & 86\% & 99\% & 3.59\% & 12\% & 80\% & 99\% \\
     & ACO & 1.98\% & 48\% & 88\% & 100\% & 17.98\% & 0\% & 1\% & 6\% & 36.69\% & 0\% & 0\% & 0\% \\
     & LLM & \textbf{ 0.14}\% & \textbf{ 96\%} & 100\% & 100\% & 0.70\% & 74\% & 100\% & 100\%  & 1.34\% & 44\% & 100\% & 100\% \\
     & Ours & 0.20\% & 95\% & \textbf{ 100\%} & \textbf{ 100\%} & \textbf{ 0.65\%} & \textbf{ 81\%} & \textbf{ 100\%} & \textbf{ 100\%} & \textbf{ 1.21\%} & \textbf{ 55\%} & \textbf{ 100\%} & \textbf{ 100\%} \\
    \hline
    \addlinespace[1pt]
    \multirow{6}{*}[+1ex]{\rotatebox[origin=c]{90}{\scriptsize{\emph{OP}}}} 
     & Tsili  & 3.85\% & 21\% & 68\% & 96\% & 9.54\% & 0\% & 2\% & 55\% & 13.80\% & 0\% & 0\% & 8\% \\
     & ACO & 3.49\% & 30\% & 76\% & 94\% & 6.24\% & 1\% & 35\% & 89\% & 7.95\% & 0\% & 15\% & 74\% \\
     & LLM & 1.47\% & 54\% & \textbf{ 95}\% & \textbf{ 99\%} & 2.04\% & 26\% & \textbf{ 96\%} & \textbf{ 100\%} & 2.10\% & 27\% & \textbf{ 96\%} & \textbf{ 99\%} \\ 
     & Ours & \textbf{ 0.59\%} & \textbf{ 69\%} & 92\% & 92\% & \textbf{ 1.08\%} & \textbf{ 54\%} & 96\% & 97\% & \textbf{ 1.67\%} & \textbf{ 34\%} & 93\% & 96\% \\
    \hline
    \addlinespace[1pt]
    \multirow{6}{*}[+1ex]{\rotatebox[origin=c]{90}{\scriptsize{\emph{CVRP}}}} &
     OR-Tools  & 3.60\% & 45\% & 69\% & 93\% & 7.87\% & 3\% & 24\% & 72\% & 8.84\% & 0\% & 15\% & 71\%  \\
     & PS  & 3.95\% & 24\% & 72\% & 93\% & 5.67\% & 2\% & 50\% & 89\% & 6.12\% & 0\% & 41\% & 89\% \\
     & LLM & \textbf{ 1.70\%} & 52\% & \textbf{ 90\%} & \textbf{ 97\%} & 4.57\% & 8\% & 59\% & 98\% & 7.24\% & 1\% & 19\% & 84\% \\
     & Ours & 1.81\% & \textbf{ 59\%} & 86\% & 95\% & \textbf{ 3.44\%} & \textbf{ 15\%} & \textbf{ 80\%} & \textbf{ 98\%} & \textbf{ 5.53\%} & 0\% & \textbf{ 45\%} & \textbf{ 89\%} \\
    \hline
    \addlinespace[1pt]
     \multirow{4}{*}[-1ex]{\rotatebox[origin=c]{90}{\scriptsize{\emph{MIS}}}} &
     Degree  & 5.89\% & 57\% & 57\% & 67\% & 7.52\% & 32\% & 42\% & 64\% & 9.61\% & 20\% & 33\% & 59\%  \\
     & Greedy  & 2.56\% & 84\% & 84\% & 86\% & 2.79\% & 67\% & 71\% & 90\% & 3.36\% & 53\% & 60\% & 83\% \\
     & LLM & 0.38\% & 97\% & 98\% & 98\% & \textbf{ 1.05}\% & 86\% & 87\% & 94\% & 2.29\% & 47\% & 57\% & 65\% \\
     & Ours & \textbf{ 0.20\%} & \textbf{ 99\%} & \textbf{ 100\%} & \textbf{ 100\%} & 1.08\% & \textbf{ 90\%} & \textbf{ 93\%} & \textbf{ 100\%} & \textbf{ 1.58\%} & \textbf{ 68\%} & \textbf{ 74\%} & \textbf{ 95\%} \\
    \hline
    \addlinespace[1pt]
    \multirow{5}{*}[-0ex]{\rotatebox[origin=c]{90}{\scriptsize{\emph{MVC}}}} 
     & Greedy   & 2.80\% & 71\% & 72\% & 89\% & 2.62\% & 44\% & 79\% & 99\% & 2.21\% & 33\% & 89\% & 100\% \\
     & Degree   & 3.53\% & 63\% & 63\% & 86\% & 2.78\% & 44\% & 78\% & 98\% & 2.63\% & 26\% & 83\% & 99\%  \\
     & LLM & 0.48\% & 93\% & 93\% & 100\% & 1.25\% & 66\% & 94\% & 100\% & 2.35\% & 38\% & 87\% & 100\% \\
     & Ours & 0.79\% & 92\% & \textbf{ 93\%} & 98\% & \textbf{ 1.07\%} & \textbf{ 66\%} & \textbf{ 98\%} & \textbf{ 100\%} & \textbf{ 1.88\%} & \textbf{ 43\%} & \textbf{ 90\%} & 95\% \\
    \hline
    \addlinespace[1pt]
    \multirow{4}{*}[-0.5ex]{\rotatebox[origin=c]{90}{\scriptsize{\emph{PFSP}}}} &
     Palmer's  & 30.52\% & 0\% & 0\% & 1\% & 30.41\% & 0\% & 0\% & 0\% & 30.68\% & 0\% & 0\% & 0\%  \\
     & NEH  & 1.33\% & 53\% & 97\% & 99\% & 2.78\% & 11\% & 90\% & 100\% & 3.56\% & 0\% & 88\% & 100\% \\
     & LLM & 0.25\% & \textbf{ 85\%} & 100\% & 100\% & \textbf{ 1.16\%} & 40\% & 100\% & 100\% & 2.62\% & 6\% & \textbf{ 99\%} & 100\% \\
     & Ours & \textbf{ 0.45\%} & 78\% & \textbf{ 100\%} & \textbf{ 100\%} & 1.28\% & \textbf{ 42\%} & \textbf{ 100\%} & \textbf{ 100\%} & \textbf{ 2.58\%} & \textbf{ 8\%} & 97\% & \textbf{ 100\%} \\
    \hline
    \addlinespace[1pt]
    \multirow{5}{*}[-0ex]{\rotatebox[origin=c]{90}{\scriptsize{\emph{JSSP}}}} &
     SPT  & 19.58\% & 2\% & 4\% & 17\% & 25.32\% & 0\% & 0\% & 1\% & 27.35\% & 0\% & 0\% & 0\% \\
     & ATC  & 20.71\% & 0\% & 12\% & 15\% & 24.30\% & 0\% & 0\% & 1\% & 27.99\% & 0\% & 0\% & 0\% \\
     & LLM & 2.86\% & 32\% & 79\% & 98\% & 9.56\% & 0\% & 8\% & 60\% & 16.25\% & 0\% & 0\% & 4\% \\
     & Ours & \textbf{ 2.31\%} & \textbf{ 38\%} & \textbf{ 87\%} & \textbf{ 100\%} & \textbf{ 6.03\%} & \textbf{ 3\%} & \textbf{ 33\%} & \textbf{ 92\%} & \textbf{ 10.28\%} & \textbf{ 0\%} & \textbf{ 3\%} & \textbf{ 43\%} \\
    \hline
\end{tabular}
\label{tab:heuristic}
\end{table*}

\section{Experiments}

\textbf{Dataset and Hyperparameters.} We fine-tune Qwen3-VL to serve as a
shared end-to-end solver for 7 CO problems: TSP, OP, CVRP, MIS, MVC, PFSP,
and JSSP, which are defined and generated in \cite{jiang2026large}. For SFT, we generate 500,000 instances for each problem, while an
additional set containing at most 3,200 instances is used for RL. 
The model is
trained for one epoch with the AdamW optimizer and BF16 precision. During
inference, greedy decoding is used for single generation, while Best-of-$N$
inference generates $N=8$ candidate solutions using stochastic decoding and selects the best solution among them. 
Following \cite{jiang2026large}, we use the same textual prompts and solvers to compute relative gaps.

\textbf{Baselines.}  We compare with general-purpose and reasoning LLMs and prompt strategies for CO including OPRO
\cite{yang2024large}, LMEA \cite{10611913}, PHP
\cite{zheng2024progressive}, and SGE \cite{iklassov2024selfguiding}.
We also compare the state-of-the-art text-only end-to-end LLM \cite{jiang2026large}, reporting SFT, SFT+RL, and
SFT+RL+Best-of-$N$ results. Following the evaluation protocol for CO, we report feasibility rate and relative gap, as defined in Section~\ref{sec:problem_statement}. The gap is computed against the optimal or best-known reference objective and is averaged over feasible solutions.


\subsection{Comparative Study}

Table~\ref{tab:model_summary_vlm} shows that fine-tuned LLM in \cite{jiang2026large} and our fine-tuned VLM outperform all the other baselines even with greedy decoding.
Visual information provides the
clearest gains on the more structurally demanding CVRP and JSSP. Compared with the fine-tuned LLM under Best-of-$N$ inference, the VLM achieves similar feasibility, and
considerably reduce the gap by $0.95\%$ and $2.35\%$, respectively. 
The VLM reduces the gap by $0.16\%$, $0.72\%$  and $0.18\%$ on TSP, OP and MVC while retaining full feasibility. Moreover, the VLM matches the LLM on PFSP and achieves higher feasibility ($97\%$ vs. $94\%$) for MIS. Therefore, our vision representation generally enhances LLM performance in CO solving, underscoring the value of visual information for developing effective end-to-end solvers.

\subsection{Comparison across Problem Scales}

We evaluate VLM across different scales, with 100 instances for each. 
Small, medium, and large instances are defined as: for scheduling problems, 5×5–10×10, 10×10–15×15, and 15×15–20×20; for other problems, 10–30, 40–60, and 70–100 nodes, respectively. We compare the VLM with the fine-tuned
text-only LLM \cite{jiang2026large} and with domain-specific baselines, which provide practical anchors for the larger instances. 
For routing, the baselines are OR-Tools \cite{ortools_routing} (TSP/CVRP),
ACO \cite{4129846} (TSP/OP), Tsiligirides
\cite{tsiligirides1984heuristic} (OP), and parallel savings (PS) (CVRP). For
graph optimization, we use the Greedy and Degree heuristics for both MIS and
MVC. For scheduling, the references are Palmer's rule
\cite{palmer1965sequencing} and NEH \cite{KALCZYNSKI200753} for PFSP, and
SPT and ATC for JSSP.
We also introduce Gap@K, the percentage of instances solved with an optimality gap below K\%. As shown in Table \ref{tab:heuristic},
our VLM performs favorably against the competing
baselines, highlighting its potential for solving CO problems at larger
scales. Its advantage becomes more pronounced and consistent as the instance size increases.

\section{Conclusion and Future Work}
\label{sec:conclusion}

We presented a VLM-based solver for CO problems. By pairing textual descriptions with
visual representations, we train VLM by SFT followed by
verifier-guided RL. The VLM improves solution quality over the
text-only LLM, with significant gain on complex CVRP and JSSP. Results across different problem scales show that the advantage of visual information becomes more pronounced as the problem size increases.
Future work will explore VLM in general optimization problems and improve its transferability in more complex CO instances.

\clearpage
\bibliographystyle{IEEEbib}
\bibliography{refs}

\end{document}